\documentclass[letterpaper, 10 pt, conference]{ieeeconf}  
\IEEEoverridecommandlockouts                              

\usepackage{amsmath} 
\usepackage{todonotes}
\usepackage{xcolor}
\usepackage{tabularx}
\usepackage{tabulary}
\usepackage{booktabs}
\usepackage{algorithm}
\usepackage[noend]{algpseudocode}
\algrenewcommand\algorithmicdo{}
\algrenewcommand\algorithmicthen{}
\usepackage[style=ieee,backend=bibtex,url=false,doi=false,isbn=false,eprint=false]{biblatex}
\AtEveryBibitem{\clearfield{issn}}
\AtEveryCitekey{\clearfield{issn}}
\usepackage[keeplastbox]{flushend}
\usepackage{graphics}
\usepackage{siunitx}
\usepackage{anyfontsize}
\usepackage{siunitx}
\usepackage[hidelinks]{hyperref}
\usepackage{cleveref}
\usepackage{subcaption}

\usepackage{tikz}
\usepackage{pgfplots}
\usetikzlibrary{arrows,decorations.pathmorphing,backgrounds,positioning,fit,petri,calc,shadows,shapes,automata,backgrounds,petri,calc,graphs,matrix}
\pgfdeclarelayer{background}
\pgfsetlayers{background,main}
\usepgfplotslibrary{groupplots}
\pgfplotsset{compat=newest}
\usepgfplotslibrary{statistics}

\makeatletter
\newcommand\fs@spaceruled{\def\@fs@cfont{\bfseries}\let\@fs@capt\floatc@ruled
  \def\@fs@pre{\vspace{5pt}\hrule height.8pt depth0pt \kern2pt}%
  \def\@fs@post{\kern2pt\hrule\relax}%
  \def\@fs@mid{\kern2pt\hrule\kern2pt}%
  \let\@fs@iftopcapt\iftrue}
\makeatother

\newcommand{\PCTAPF}{PC-TAPF}

\newcommand{\numrobots}{n}

\newcommand{\numtasks}{m}

\newcommand{\RobotPosition}[2][]{x_{#1}^{#2}}
\newcommand{\RobotState}[2][]{\RobotPosition[#1]{#2}}

\newcommand{\ObjectState}[2][]{o_{#1}^{#2}}

\newcommand{\FinalTime}{T}
\newcommand{\StartTime}[1][]{t_{#1}^{0}}
\newcommand{\CompletionTime}[1][]{t_{#1}^{F}}

\newcommand{\TaskStartTime}[1][]{\tau_{#1}^{0}}
\newcommand{\TaskCompletionTime}[1][]{\tau_{#1}^{F}}
\newcommand{\RobotStartTime}[1][]{t_{#1}^{0}}

\newcommand{\PickupLocation}[1][]{s_{#1}}
\newcommand{\DropoffLocation}[1][]{g_{#1}}

\newcommand{\distmatrix}[2][]{d({#1},{#2})}
\newcommand{\assignment}[2][]{A_{#1#2}}
\newcommand{\predecessors}[1]{\textsc{pred}(#1)}
\newcommand{\successors}[1]{\textsc{succ}(#1)}

\newcommand{\ProjectSpec}{S}
\newcommand{\ProjectGraphVertices}{\mathcal{V}}
\newcommand{\ProjectGraphEdges}{\mathcal{E}}

\newcommand{\ProjectSchedule}{G}

\newcommand{\ScheduleVertices}{\mathcal{V}}
\newcommand{\ScheduleEdges}{\mathcal{E}}

\newcommand{\Vertices}{\textsc{vertices}}
\newcommand{\ScheduleNode}{v}
\newcommand{\vtx}{v}

\newcommand{\vtxTWO}{v^{\prime}}

\newcommand{\ROBOTat}{\textsc{robot\_at}}
\newcommand{\OBJECTat}{\textsc{object\_at}}
\newcommand{\GOaction}{\textsc{go}}
\newcommand{\COLLECTaction}{\textsc{collect}}
\newcommand{\CARRYaction}{\textsc{carry}}
\newcommand{\DEPOSITaction}{\textsc{deposit}}
\newcommand{\OPERATIONvtx}{\textsc{operation}}

\newcommand{\ROBOTatAbbrev}[1]{\textsc{r#1}}
\newcommand{\OBJECTatAbbrev}[1]{\textsc{o#1}}
\newcommand{\OPERATIONvtxAbbrev}[1]{\textsc{op{#1}}}

\newcommand{\RobotID}{i}
\newcommand{\RobotIDtwo}{j}
\newcommand{\ObjectID}{j}

\newcommand{\processtime}[1][]{\Delta t_{#1}}

\newcommand{\pctapfSolution}{S}

\newcommand{\RoutePlan}{P}

\newcommand{\AgentPath}[1]{p_{#1}}
\newcommand{\solutionCost}{T}

\newcommand{\milp}{\textsc{milp}}

\newcommand{\LevelOne}{NBS}
\newcommand{\LevelOneLong}{Sequential Next-Best Assignment Search}

\newcommand{\bestCost}{\solutionCost^\star} 
\newcommand{\bestSolution}{\pctapfSolution^\star}
\newcommand{\nullSolution}{nothing}
\newcommand{\lowerBoundCost}{\underline{\solutionCost}}

\newcommand{\problemInstance}{\textsc{pc-tapf}}
\newcommand{\formulateMILP}{\textsc{formulate\_milp}}

\newcommand{\solveMILP}{\textsc{solve}}
\newcommand{\constraints}{constraints}

\newcommand{\LevelTwo}{CBS}
\newcommand{\LevelTwoLong}{Conflict-Based Search}
\newcommand{\CBSStateConstraint}{c_S}

\newcommand{\CBSActionConstraint}{c_A}
\newcommand{\CBSStateConstraintDef}[3][]{(#1,#2,#3)}
\newcommand{\CBSActionConstraintDef}[3][]{(#1,#2,#3)}

\newcommand{\LevelThree}{ISPS}
\newcommand{\LevelThreeLong}{Incremental Slack-Prioritized Search}

\newcommand{\Slack}{\textsc{slack}}

\newcommand{\ClosedSet}{\mathcal{C}}

\newcommand{\NodeQueue}{Q}

\newcommand{\addPathToRoutePlan}{\textsc{add\_path}}

\newcommand{\LevelFour}{A$^\star_{SC}$}
\newcommand{\LevelFourLong}{Slack-and-Collision-aware Tie-breaking A$^\star$}

\newcommand{\PathCost}{\mathbf{c}} 
\newcommand{\HeuristicCost}{h}

\newcommand{\CountConflicts}{\textsc{count\_conflicts}}

\newcommand{\pathstate}{s}

\newcommand{\numproblems}{384}

\newcommand{\reals}{{\mbox{\bf R}}}

\newcommand{\booleans}{{\mbox{\bf B}}}

\title{\LARGE \bf
Optimal Sequential Task Assignment and Path Finding \\ for Multi-Agent Robotic Assembly Planning
}

\author{Kyle Brown \and Oriana Peltzer \and Martin A. Sehr \and Mac Schwager \and Mykel J. Kochenderfer%
\thanks{*This work was supported by Siemens and the National Science Foundation.}%
\thanks{$^{1}$Kyle Brown and Oriana Peltzer are with the Stanford University Department of Mechanical Engineering, {\tt\small \{kjbrown7, peltzer\}@stanford.edu}.}%
\thanks{$^{2}$Martin A. Sehr is with Corporate Technology, Siemens Corporation, Berkeley, CA 94704, USA, {\tt\small martin.sehr@siemens.com}.}%
\thanks{$^{3}$Mac Schwager and Mykel Kochenderfer are with the Stanford University Department of Aeronautics and Astronautics,{\tt\small \{mykel, schwager \}@stanford.edu}.}%
}

\newcommand{\graphicscale}{0.8}
\newcommand{\nodescale}{1.0}

\begin{document}

\tikzset{level plate/.style={%
      thick,
      fill=black!10,
      text depth=0pt,
      rounded corners,
      inner ysep=4pt,
      inner xsep=4pt,
  },
  empty level plate/.style={%
      text depth=0pt,
      anchor=west,
      align=center,
      minimum width=0cm,
      minimum height=0cm,
      inner ysep=0pt,
      inner xsep=0pt
  },
  data block/.style={%
      rounded corners,
      text depth=0pt,
      thick,
      minimum width=0cm,
      minimum height=0cm,
      fill=blue!35,
      thick,
      inner ysep=5pt,
      inner xsep=5pt
  },
  empty data block/.style={%
      minimum height=0cm,
      minimum width=0cm,
      text depth=0pt,
      inner ysep=0pt,
      inner xsep=0pt
  },
  function block/.style={%
      thick,
      minimum width=0cm,
      minimum height=0cm,
      text depth=0pt,
      fill=red!40,
      rounded corners,
      inner ysep=5pt,
      inner xsep=5pt,
      minimum width=0.5cm
  },
  empty function block/.style={%
      minimum width=0cm,
      minimum height=0cm,
      text depth=0pt,
      rounded corners,
      inner sep=0pt
  },
  intermediate output block/.style={%
      minimum width=0cm,
      minimum height=0cm,
      text depth=0pt,
      rounded corners,
      inner sep=0pt
  },
  thick edge/.style={%
    >=latex,
    ->,
    thick,
    shorten >=1pt,
    shorten <=1pt,
  }
}
\tikzset{BaseScheduleNode/.style={inner sep=3pt, outer sep=1pt, align=center,minimum size=0pt}}
\tikzset{TaskNode/.style={BaseScheduleNode,shape=circle, text=black!, fill=black!20, opacity=1}}
\tikzset{RobotNode/.style={BaseScheduleNode,shape=circle, text=black!, fill=green!20, opacity=1}}
\tikzset{RobotTerminalNode/.style={BaseScheduleNode,shape=diamond, text=black!, fill=green!20, opacity=1}}
\tikzset{ObjectNode/.style={BaseScheduleNode,shape=circle, text=black!, fill=orange!20, opacity=1}}
\tikzset{ObjectTerminalNode/.style={BaseScheduleNode,shape=diamond, text=black!, fill=orange!20, opacity=1}}
\tikzset{ActionNode/.style={BaseScheduleNode,shape=rectangle, rounded corners=4pt,text=black!, fill=blue!20, opacity=1}}
\tikzset{OpNode/.style={BaseScheduleNode,shape=regular polygon,regular polygon sides=4, rounded corners=0pt, text=black!, fill=red!20, opacity=1}}
\tikzset{ScheduleEdge/.style={<-,>=latex, shorten < = 1pt, shorten > = 1pt,draw,thick}}
\tikzset{BlockArrow/.style={single arrow, fill=red!20, minimum height=1.5em,minimum width=1.5em,
    single arrow head extend=0.15cm, outer sep=0pt}}

\newcommand{\LevelOneColor}{black}
\newcommand{\LevelTwoColor}{black}
\newcommand{\LevelThreeColor}{black}
\newcommand{\LevelFourColor}{black}
\newcounter{overviewcounter}
\setcounter{overviewcounter}{1}
\newcommand{\compoverviewcounter}[1]{\ifthenelse{\value{overviewcounter}>#1}{black}{white}}

\newcounter{milpEdgesDrawingMode}
\setcounter{milpEdgesDrawingMode}{1}

\newcommand{\overrideedgecolor}{black}

\newcommand{\stackwidth}{70pt}
\newcommand{\stackheight}{20pt}

\newcommand\Wider[2][3em]{%
\makebox[\linewidth][c]{%
  \begin{minipage}{\dimexpr\textwidth+#1\relax}
  \raggedright#2
  \end{minipage}%
  }%
}

\maketitle
\thispagestyle{empty}
\pagestyle{empty}

\newcommand{\kylenote}{\textcolor{red}}
\begin{abstract}

  We study the problem of sequential task assignment and collision-free routing for large teams of robots in applications with inter-task precedence constraints (e.g., task $A$ and task $B$ must both be completed before task $C$ may begin). Such problems commonly occur in assembly planning for robotic manufacturing applications, in which sub-assemblies must be completed before they can be combined to form the final product. We propose a hierarchical algorithm for computing makespan-optimal solutions to the problem. The algorithm is evaluated on a set of randomly generated problem instances where robots must transport objects between stations in a ``factory'' grid world environment. In addition, we demonstrate in high-fidelity simulation that the output of our algorithm can be used to generate collision-free trajectories for non-holonomic differential-drive robots.
  
  \end{abstract}
  
  \section{Introduction}
  Consider a factory environment with an array of manufacturing stations and a fleet of autonomous mobile robots. The \emph{project} consists of an initial set of objects and a prescribed set of manufacturing operations that will incrementally transform those initial objects into a final object. The goal is to assign and route robots to transport objects between stations so that the project makespan---the time from start to completion---is minimized. 
  We call this problem precedence-constrained multi-agent task assignment and path-finding (\PCTAPF).
  
  \PCTAPF{} generalizes the multi agent pathfinding (MAPF) problem with task assignment, sometimes called the anonymous MAPF \cite{Yu2013}. Whereas anonymous MAPF involves a \emph{one off} task assignment problem (each robot performs no more than one task) in which all tasks are independent, \PCTAPF{} involves \emph{sequential} task assignment (each robot may be assigned to a sequence of tasks) and incorporates temporal precedence constraints between tasks (e.g., task $A$ and task $B$ must both be completed before task $C$ may begin). Both problems involve collision-free routing, but the \PCTAPF{} routing problem is affected by the same precedence constraints that are present in the assignment problem.
  Although minimum-makespan anonymous MAPF problems with homogeneous agents can be solved in polynomial time \cite{Yu2013}, the \PCTAPF{} problem is NP-hard. The difference lies in the ``cross-schedule dependencies'' that arise from the inter-task precedence constraints \cite{Korsah}.
    As the ratio of tasks to agents increases, the sequential assignment problem approaches a traveling salesman problem.
    As the breadth of the task dependency graph (i.e., the degree to which it is parallelizable) increases, the route-planning problem becomes increasingly congested and potentially ill-matched with optimistic task assignment.
      However, \PCTAPF{} problems often exhibit useful structure that can be exploited to efficiently find optimal solutions.
  
  
  Various methods have been proposed for solving problems that are mathematically similar to \PCTAPF{}.
      Some perform joint task assignment and collision-free route-planning, but do not handle dependencies between tasks \cite{Honig2018,Wagner2012,Ma2016b,Yu2015a,Ma2018a}.
    Others handle tasks with temporal dependencies, but ignore the routing problem or assume that agents will never collide \cite{Dohn2011,Bredstrom2011,Bredstrom2008}.
          To the best of our knowledge, no solver has been proposed to optimally solve the combined sequential task assignment and routing problems with inter-task precedence constraints and collision-free constraints.
  
  
  In this article, we propose a hierarchical algorithm for optimally solving \PCTAPF{} problems.
      The first level assigns tasks to robots by solving a relaxed problem that ignores non-collision constraints.
          The task assignments are passed down to a Conflict-Based Search (CBS) level, which searches over a constraint tree for an optimal collision-free set of paths \cite{Sharon2012}. At each node of the constraint tree, CBS calls a lower level routine that incrementally constructs a joint route plan by iterating over a dependency graph and repeatedly calling a modified version of A$^\star$ \cite{Hart1968}.
              We evaluate the runtime of the algorithm and its component parts on a suite of problem instances that vary in size and structure, and show that even large problem instances (i.e. 40 robots, 60 tasks) can be efficiently solved to optimality in most cases.
                  We also demonstrate our solver with a fleet of non-holonomic differential-drive robots in high-fidelity simulation.
  
  \section{Background}
      
  Many variants of multi-agent pathfinding (MAPF) \cite{Silver2005} problems exist in the literature. Related problems include MAPF with Deadlines (MAPF-DL) \cite{Ma2018a}, multi-agent pick-up and delivery (MAPD) \cite{Ma2017c}, and combined target-assignment and path-finding (TAPF) with teams of heterogeneous agents \cite{Yu2016,Ma2016b}.
  Optimal solvers for path-finding problems are often based on Conflict-Based Search (CBS) \cite{Sharon2012}, Network flow \cite{Yu2016} or subdimensional expansion \cite{Wagner2012}.
  Conflict-Based Search with Task Assignment (CBS-TA) addresses heterogeneous anonymous MAPF problems \cite{Honig2018}. CBS-TA searches over a forest of CBS constraint trees, where each tree in the search forest corresponds to a different task assignment matrix. Our algorithm is similar, in the sense that it solves a relaxed ``assignment'' problem at the top level and queries a lower-level route-planner, iterating through all possible assignments in best-first order until there is zero gap between the lower bound from the assignment problem and the best route plan.
   
          
  Several types of vehicle routing and scheduling problems (VRSPs) incorporate temporal dependencies---such as precedence, synchronization, and time-windowing constraints---between tasks.
  \citeauthor{Bredstrom2008} introduce a traveling repairman VRSP with precedence and synchronization constraints \cite{Bredstrom2008}.
      \citeauthor{Dohn2011} introduce a formulation for modeling general temporal dependencies between tasks \cite{Dohn2011}.
  Exact methods for VRSPs with temporal dependencies usually involve solving a mixed integer linear program (MILP) \cite{Dohn2011}. We do the same at the top level of our hierarchical algorithm. Because such solution methods are often intractable for large problem instances, many decentralized approaches accept a degree of suboptimality in exchange for savings in computation time. These include algorithms like M+, wherein robots negotiate with each other in a ``task market’’ \cite{Botelho}. 
  
  
  
  \section{Problem Statement}
  
\begin{figure}
  \centering
    \centering
    \vspace{5pt}
    \includegraphics[width=0.55\linewidth]{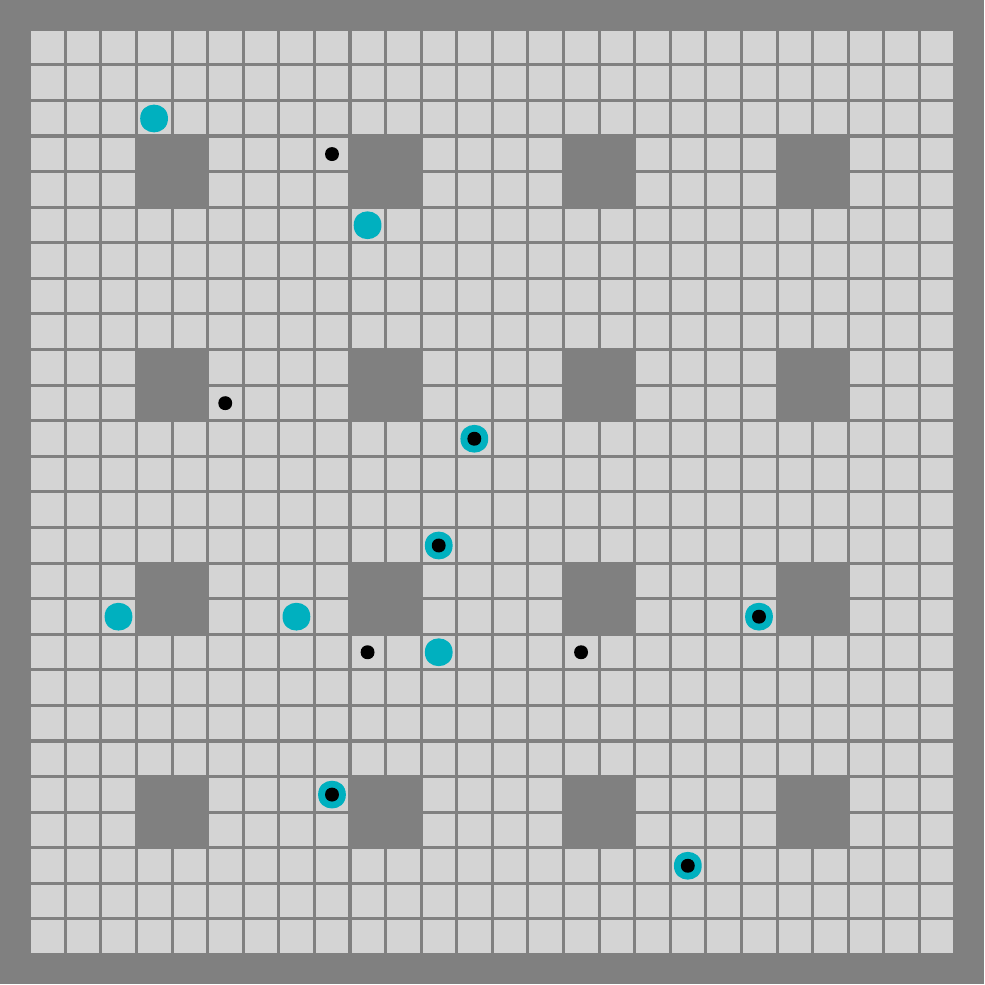}
    \caption{The factory environment. Dark gray regions represent obstacles (manufacturing stations). Blue disks represent robots, and smaller black disks represent objects.}
    \label{fig:factory_env}
  \end{figure}

  The factory environment is modeled as a two-dimensional grid world. Manufacturing stations are regularly spaced throughout the environment, and each station is surrounded by an array of pick-up and drop-off zones, at which objects may be collected and deposited, respectively. The discrete state $\RobotState[\RobotID]{t}$ of robot $\RobotID$ at time $t$ corresponds to a grid cell in the environment. At each time step, each robot may remain in place or move in any of the compass directions to arrive in an adjacent grid cell at the next time step. Cells corresponding to manufacturing stations may not be entered. robot $\RobotID$ may collect object $\ObjectID$ if they occupy the same grid cell (i.e., $\RobotState[\RobotID]{t} = \ObjectState[\ObjectID]{t}$, where $\ObjectState[\ObjectID]{t}$ denotes the object state), and the object moves with the robot until the robot deposits it. The environment is shown in \cref{fig:factory_env}.
  
  A \emph{project specification} $\ProjectSpec$ defines the set of manufacturing operation that must be performed. Each operation consumes one or more input objects and produces (except in the case of the no-output terminal operation) a single output object. Precedence constraints arise whenever the output of one operation is required as an input to another operation. The project specification also defines the pick-up and drop-off locations of all associated objects. \Cref{fig:proj_spec} depicts a simple project specification. The project \emph{makespan} is defined as the time from start ($t=0$) to completion of the terminal operation.

  
  
  
  \begin{table}[t]
    \centering
    \renewcommand{\arraystretch}{1.2}
    \hspace*{-0.4cm}
    \caption{Notation\label{tab:notation}}
    \begin{tabular*}{0.49\linewidth}{@{} l@{\extracolsep{\fill}} l@{ }}
       \toprule
        $\numrobots$ & number of robots \\
        $\numtasks$ & number of objects \\
        $\RobotState[\RobotID]{}$ & state of robot $\RobotID$ \\
        $\ObjectState[\ObjectID]{}$ & state of object $\ObjectID$ \\
        $\AgentPath{\RobotID}=\RobotPosition[\RobotID]{0:t}$ & trajectory of agent $\RobotID$ \\
        & \\
        \bottomrule
    \end{tabular*}%
    \hspace*{0.02\linewidth}%
    \begin{tabular*}{0.49\linewidth}{@{} l@{\extracolsep{\fill}} l@{ }}
       \toprule
        $\ProjectSchedule = (\ScheduleVertices, \ScheduleEdges)$ & operating schedule \\
        $\ScheduleNode \in \ScheduleVertices$ & schedule vertex \\
        $(\vtx \rightarrow \vtxTWO) \in \ProjectGraphEdges$ & schedule edge \\
        $\vtx.\StartTime[]$ &  vertex start time \\
        $\vtx.\processtime[]$ & vertex duration \\
        $\vtx.\CompletionTime[]$ & vertex end time \\
        \bottomrule
    \end{tabular*}
  \end{table}

  \subsection{Operating Schedules and Route Plans}
  
  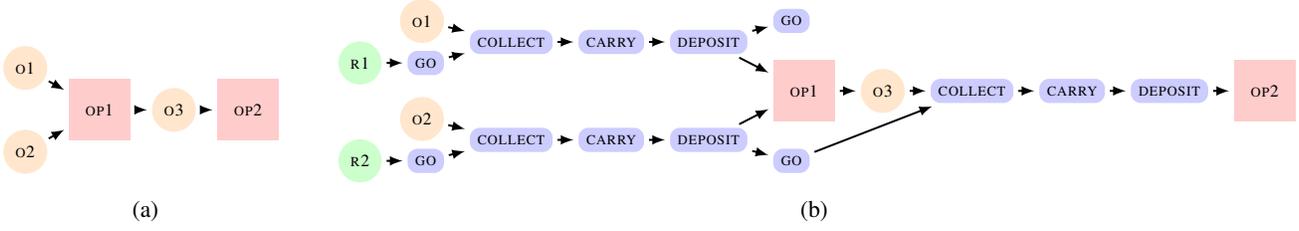
\begin{figure*}
    \vspace{5pt}
    \centering
    \begin{subfigure}[b]{0.25\linewidth}
      \centering
      \begin{tikzpicture}[remember picture]
        \node[] (a) {\resizebox{0.85\textwidth}{!}{
\begin{tikzpicture}[scale=\graphicscale,
  every node/.style={scale=\nodescale}]

\node [OpNode,font=\scriptsize] (op2) {$\OPERATIONvtxAbbrev{2}$} [ScheduleEdge,anchor=east,grow=left,growth parent anchor=west,level distance=3mm,font=\scriptsize]
  child { node [ObjectNode] (o3) {$\OBJECTatAbbrev{3}$}
    child { node [OpNode] (op1) {$\OPERATIONvtxAbbrev{1}$} [sibling distance=15mm]
      child { node [ObjectNode] (o1) {$\OBJECTatAbbrev{1}$} }
      child { node [ObjectNode] (o2) {$\OBJECTatAbbrev{2}$} }
    }
  };

\end{tikzpicture}}};
      \end{tikzpicture}
      \caption{}
      \label{fig:proj_spec}
    \end{subfigure}%
    \begin{subfigure}[b]{0.75\linewidth}
      \begin{tikzpicture}[remember picture]
        \newcommand{\freezeColor}{black!10}
        \newcounter{splitFlag}
        \newcounter{freezeFlag}
        \newcounter{mergeSchedule}

        \newcommand{\stepSep}{30mm}
        \newcommand{\figwidth}{0.98\textwidth}

        \setcounter{splitFlag}{0}
        \setcounter{freezeFlag}{0}
        \setcounter{mergeSchedule}{0}
        \setcounter{milpEdgesDrawingMode}{1}
        \renewcommand{\overrideedgecolor}{black}
        \node[] (a) {};
        \node[inner sep=0pt,outer sep=0pt] at (a) {\resizebox{\figwidth}{!}{
\newcommand{\sibdistONE}{14mm}
\newcommand{\sibdistTWO}{6mm}
\newcommand{\leveldist}{3mm}
\begin{tikzpicture}[font=\scriptsize]

\node [OpNode,font=\scriptsize] (op2) {$\OPERATIONvtxAbbrev{2}$} [ScheduleEdge,anchor=east,grow=left,growth parent anchor=west,level distance=\leveldist,sibling distance=\sibdistTWO]
  child { node [ActionNode] (deposit3) {$\DEPOSITaction$} 
  child { node [ActionNode] (carry3) {$\CARRYaction$}
    child { node [ActionNode] (collect3) {$\COLLECTaction$} 
    child { node [ObjectNode] (o3) {$\OBJECTatAbbrev{3}$}
      child { node [OpNode] (op1) {$\OPERATIONvtxAbbrev{1}$} [sibling distance=\sibdistONE]
      child { node [ActionNode] (deposit1) {$\DEPOSITaction$} 
        child { node [ActionNode] (carry1) {$\CARRYaction$} 
        child { node [ActionNode] (collect1) {$\COLLECTaction$} [sibling distance=\sibdistTWO]
          child { node [ObjectNode] (o1) {$\OBJECTatAbbrev{1}$} }
          child { node [ActionNode] (go1) {$\GOaction$} 
          child { node [RobotNode] (r1) {$\ROBOTatAbbrev{1}$} }
          edge from parent[draw=none] 
          }
        }
        }
      }
      child { node [ActionNode] (deposit2) {$\DEPOSITaction$} 
        child { node [ActionNode] (carry2) {$\CARRYaction$} 
        child { node [ActionNode] (collect2) {$\COLLECTaction$} [sibling distance=\sibdistTWO]
          child { node [ObjectNode] (o2) {$\OBJECTatAbbrev{2}$} }
          child { node [ActionNode] (go2) {$\GOaction$} 
          child { node [RobotNode] (r2) {$\ROBOTatAbbrev{2}$} }edge from parent[draw=none]
          }
        }
        }
      }
      }
    }
    }
  }
  };

  \ifthenelse{\value{mergeSchedule}=1}{
    \node [OpNode,below=15mm of op2, font=\scriptsize] (op4) {$\OPERATIONvtxAbbrev{4}$} [ScheduleEdge,anchor=east,grow=left,growth parent anchor=west,level distance=\leveldist,sibling distance=\sibdistTWO]
  child { node [ActionNode] (deposit6) {$\DEPOSITaction$} 
  child { node [ActionNode] (carry6) {$\CARRYaction$}
    child { node [ActionNode] (collect6) {$\COLLECTaction$} 
    child { node [ObjectNode] (o6) {$\OBJECTatAbbrev{6}$}
      child { node [OpNode] (op3) {$\OPERATIONvtxAbbrev{3}$} [sibling distance=\sibdistONE]
      child { node [ActionNode] (deposit4) {$\DEPOSITaction$} 
        child { node [ActionNode] (carry4) {$\CARRYaction$} 
          child { node [ActionNode] (collect4) {$\COLLECTaction$} 
            child { node [ObjectNode] (o4) {$\OBJECTatAbbrev{4}$} }
            }
          }
        }
      child { node [ActionNode] (deposit5) {$\DEPOSITaction$} 
        child { node [ActionNode] (carry5) {$\CARRYaction$} 
          child { node [ActionNode] (collect5) {$\COLLECTaction$}
            child { node [ObjectNode] (o5) {$\OBJECTatAbbrev{5}$} }
          }
        }
      }
      }
    }
    }
  }
  };
  }{}

  \ifthenelse{\value{splitFlag}=1} {
    \node [ActionNode,right=\leveldist of deposit1,yshift=3mm] (go5) {$\GOaction$};
    \draw[ScheduleEdge,->] (deposit1) -- (go5);
    \node [ActionNode,right=\leveldist of go5] (go3) {$\GOaction$};
    \draw[ScheduleEdge,->] (go5) -- (go3);

    \node [ActionNode,right=\leveldist of deposit2,yshift=-3mm] (go6) {$\GOaction$};
    \draw[ScheduleEdge,->] (deposit2) -- (go6);
    \node [ActionNode,right=\leveldist of go6] (go4) {$\GOaction$};
    \draw[ScheduleEdge,->] (go6) -- (go4);

    \ifthenelse{\value{freezeFlag}=1} {
      \node [ActionNode,fill=\freezeColor] at (go5) {$\GOaction$};
      \node [ActionNode,fill=\freezeColor] at (go6) {$\GOaction$};
    }{}
  }{%
    \node [ActionNode,right=\leveldist of deposit1,yshift=3mm] (go3) {$\GOaction$};
    \draw[ScheduleEdge,->] (deposit1) -- (go3);
    \node [ActionNode,right=\leveldist of deposit2,yshift=-3mm] (go4) {$\GOaction$};
    \draw[ScheduleEdge,->] (deposit2) -- (go4);
  }
  
  \ifthenelse{\value{milpEdgesDrawingMode}=1} {
    \draw[ScheduleEdge,->,draw=\overrideedgecolor] (go1) -- (collect1);
    \draw[ScheduleEdge,->,draw=\overrideedgecolor] (go2) -- (collect2);
    \draw[ScheduleEdge,->,draw=\overrideedgecolor] (go4) -- (collect3);
  }{}
  \ifthenelse{\value{milpEdgesDrawingMode}=2} {
    \draw[ScheduleEdge,->,draw=\overrideedgecolor] (go1) -- (collect1);
    \draw[ScheduleEdge,->,draw=\overrideedgecolor] (go2) -- (collect2);
  }{}
  \ifthenelse{\value{milpEdgesDrawingMode}=3} {
    \draw[ScheduleEdge,->,draw=\overrideedgecolor] (go1) -- (collect1);
    \draw[ScheduleEdge,->,draw=\overrideedgecolor] (go2) -- (collect2);
    \draw[ScheduleEdge,->,draw=\overrideedgecolor] (go3) -- (collect3);
  }{}

  \ifthenelse{\value{freezeFlag}=1} {
    \node [OpNode,fill=\freezeColor] at (op1) {$\OPERATIONvtxAbbrev{1}$}; 
    \node [ActionNode,fill=\freezeColor] at (deposit1) {$\DEPOSITaction$}; 
    \node [ActionNode,fill=\freezeColor] at (carry1) {$\CARRYaction$};
    \node [ActionNode,fill=\freezeColor] at (collect1) {$\COLLECTaction$}; 
    \node [ObjectNode,fill=\freezeColor] at (o1) {$\OBJECTatAbbrev{1}$}; 
    \node [ActionNode,fill=\freezeColor] at (go1) {$\GOaction$}; 
    \node [RobotNode,fill=\freezeColor] at (r1) {$\ROBOTatAbbrev{1}$};
    \node [ActionNode,fill=\freezeColor] at (deposit2) {$\DEPOSITaction$};
    \node [ActionNode,fill=\freezeColor] at (carry2) {$\CARRYaction$};
    \node [ActionNode,fill=\freezeColor] at (collect2) {$\COLLECTaction$};
    \node [ObjectNode,fill=\freezeColor] at (o2) {$\OBJECTatAbbrev{2}$};
    \node [ActionNode,fill=\freezeColor] at (go2) {$\GOaction$};
    \node [RobotNode,fill=\freezeColor] at (r2) {$\ROBOTatAbbrev{2}$};
  }{}

\end{tikzpicture}
}};
      \end{tikzpicture}
      \caption{}
      \label{fig:proj_schedule}
    \end{subfigure}
    \caption{(\protect\subref{fig:proj_spec}) An example project specification and (\protect\subref{fig:proj_schedule}) a possible corresponding schedule graph with $\numrobots=2$ robots. robot $1$ is assigned to deliver object $1$, and robot $2$ is assigned to first deliver object $2$ and then deliver object $3$ once Operation $1$ is complete.}
    \label{fig:project_schedule_graph}
  \end{figure*}

  The \emph{operating schedule} $\ProjectSchedule = (\ProjectGraphVertices,\ProjectGraphEdges)$ is a directed acyclic graph (DAG). Each vertex $\vtx \in \ProjectGraphVertices$ of the graph corresponds to one of the following discrete high-level events or activities: initial object condition $\OBJECTat{}$, initial robot condition $\ROBOTat{}$, manufacturing operation $\OPERATIONvtx{}$, and the robot actions $\GOaction{}$, $\COLLECTaction{}$, $\CARRYaction{}$ and $\DEPOSITaction{}$. An edge $(\vtx \rightarrow \vtxTWO) \in \ProjectGraphEdges$ denotes a precedence constraint, meaning that the activity associated with $\vtxTWO$ may not begin until the activity associated with $\vtx$ has been completed. 
  The required topological structure of the schedule is intuitive: In order to perform a transport task, a robot must \emph{go} to the appropriate pick-up location, \emph{collect} an object, \emph{carry} it through the factory, and finally \emph{deposit} it at the drop-off location. A manufacturing operation may only begin once all input objects have been deposited at the prescribed drop-off points. An output object becomes available for collection only once the associated operation has been completed. 
  An example operating schedule is shown in \cref{fig:proj_schedule}.

  Each vertex $\vtx$ has a start time $\vtx.\StartTime[]$ and completion time $\vtx.\CompletionTime[]$. For root nodes only (i.e., $\ROBOTat{}$ vertices and the initial $\OBJECTat{}$ vertices), start times are fixed. The completion time of every vertex is equal to $\vtx.\StartTime[] + \vtx.\processtime[]$, where $\processtime[]$ is the duration of the activity.
    For $\ROBOTat$ and $\OBJECTat$ nodes, $\processtime[] = 0$ by definition. Each $\COLLECTaction$, $\DEPOSITaction$ and $\OPERATIONvtx$ node may have a fixed positive $\processtime[]$. For navigation nodes ($\GOaction$ and $\CARRYaction$), $\processtime[]$ is a variable that depends on how quickly the robot travels from the node's start location to its destination location. 
  

  An operating schedule is \emph{valid} if and only if (a) all vertices have the correct number of incoming and outgoing edges to the appropriate neighbor vertices and (b) there are no cycles in the graph. An example of a cycle would be if robot $\RobotID$ were assigned to first deliver object $\ObjectID$ and then deliver one of object $\ObjectID$'s prerequisites, which would require traveling backward through time. Constructing a valid schedule corresponds to solving a sequential task assignment problem. 

  A \emph{route plan} is denoted by $\RoutePlan = (\AgentPath{1}, \ldots, \AgentPath{\numrobots})$, where $\AgentPath{\RobotID} = (\RobotPosition[\RobotID]{0}, \RobotPosition[\RobotID]{1}, \ldots, \RobotPosition[\RobotID]{t})=\RobotPosition[\RobotID]{0:t}$ is the trajectory of agent $\RobotID$ from time $0$ to $t$. A route plan is \emph{consistent} if and only if it satisfies all environment constraints in addition to the constraints associated with the operating schedule (i.e., each robot must be at the relevant pick-up and drop-off locations at the time steps specified by the operating schedule). These latter constraints may be understood as boundary conditions that partition an individual robot’s trajectory into disjoint segments. A route plan is \emph{valid} if and only if it is consistent and it contains no \emph{conflicts} between trajectories. A \emph{state-conflict} between agents $\RobotID$ and $\RobotIDtwo$ is said to occur at time $t$ if $\RobotState[\RobotID]{t} = \RobotState[\RobotIDtwo]{t}$. An \emph{action conflict} occurs if $(\RobotState[\RobotID]{t} = \RobotState[\RobotIDtwo]{t+1}) \land (\RobotState[\RobotID]{t+1} = \RobotState[\RobotIDtwo]{t})$, where the two agents switch places in a single time step (which would require that they pass through each other).
  
  

  A \emph{solution} consists of a operating schedule and corresponding route plan. A solution is valid if and only if the schedule and the route plan are both valid. A valid solution is optimal if and only if no other valid solution has a lower \emph{makespan}. The project makespan $\FinalTime$ is defined as the total number of timesteps from the beginning of a project ($t = 0$) until the terminal operation is completed. 
  
  \section{Methods}\label{sec:methods}

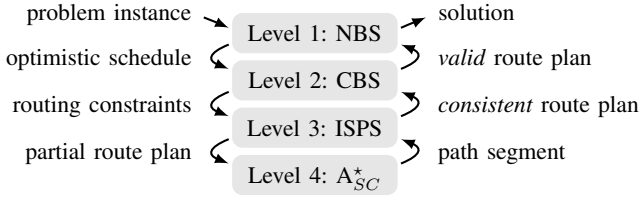
\begin{figure}
    \vspace{5pt}
    \centering
    {
      \renewcommand{\nodescale}{0.9}
      \renewcommand{\stackwidth}{70pt}
      \renewcommand{\stackheight}{17pt}
      \resizebox{0.48\textwidth}{!}{
\begin{tikzpicture}[scale=\graphicscale, every node/.style={scale=\nodescale}]
  \node[level plate, minimum height=\stackheight, minimum width=\stackwidth,text=\LevelOneColor] (level1) {Level 1: \LevelOne{}};%
  \node[level plate, minimum height=\stackheight, minimum width=\stackwidth, below = 2pt of level1,text=\LevelTwoColor] (level2) {Level 2: \LevelTwo{}};
  \node[level plate, minimum height=\stackheight, minimum width=\stackwidth, below = 2pt of level2,text=\LevelThreeColor] (level3) {Level 3: \LevelThree{}};
  \node[level plate, minimum height=\stackheight, minimum width=\stackwidth, below = 2pt of level3,text=\LevelFourColor] (level4) {Level 4: \LevelFour{}};

  \newcommand{\labelxshift}{15pt}
  \node[intermediate output block, left = \labelxshift of level1.north west,yshift=0pt] (problem instance label) {problem instance};
  \node[intermediate output block, align=center,right = \labelxshift of level1.north east,yshift=0pt] (optimal solution label) {solution};
  \newcommand{\ctrlangle}{24}
  \newcommand{\ctrlmag}{5pt}
  \draw[thick edge,shorten <=5pt] (problem instance label.east)  .. controls +(-\ctrlangle:\ctrlmag) and +(-\ctrlangle:-\ctrlmag) .. ([yshift=2pt] level1.west);
  \draw[thick edge,<-,shorten <=5pt] (optimal solution label.west)  .. controls +(\ctrlangle:-\ctrlmag) and +(\ctrlangle:\ctrlmag) .. ([yshift=2pt] level1.east);
  \renewcommand{\ctrlangle}{22}
  \renewcommand{\ctrlmag}{12pt}
  \draw[thick edge] ([yshift=5pt] level2.east)  .. controls +(\ctrlangle:\ctrlmag) and +(-\ctrlangle:\ctrlmag) .. node[intermediate output block,pos=0.5,right=8pt]{\emph{valid} route plan} ([yshift=-5pt] level1.east);
  \draw[thick edge,<-] ([yshift=5pt] level2.west)  .. controls +(-\ctrlangle:-\ctrlmag) and +(\ctrlangle:-\ctrlmag) .. node[intermediate output block,pos=0.5,left=8pt,align=right]{optimistic schedule} ([yshift=-5pt] level1.west);
  \draw[thick edge] ([yshift=5pt] level3.east)  .. controls +(\ctrlangle:\ctrlmag) and +(-\ctrlangle:\ctrlmag) .. node[intermediate output block,pos=0.5,right=8pt]{\emph{consistent} route plan} ([yshift=-5pt] level2.east);
  \draw[thick edge,<-] ([yshift=5pt] level3.west)  .. controls +(-\ctrlangle:-\ctrlmag) and +(\ctrlangle:-\ctrlmag) .. node[intermediate output block,pos=0.5,left=8pt,align=right]{routing constraints} ([yshift=-5pt] level2.west);
  \draw[thick edge] ([yshift=5pt] level4.east)  .. controls +(\ctrlangle:\ctrlmag) and +(-\ctrlangle:\ctrlmag) .. node[intermediate output block,pos=0.5,right=8pt]{path segment} ([yshift=-5pt] level3.east);
  \draw[thick edge,<-] ([yshift=5pt] level4.west)  .. controls +(-\ctrlangle:-\ctrlmag) and +(\ctrlangle:-\ctrlmag) .. node[intermediate output block,pos=0.5,left=8pt,align=right]{partial route plan} ([yshift=-5pt] level3.west);
\end{tikzpicture}}
    }
    \caption{A schematic overview of our algorithm.}\label{fig:solver_overview}
  \end{figure}

  We propose a four-level hierarchical planning algorithm to optimally solve \PCTAPF{} problems. 
  Each of the four levels is described in this section. A graphical summary of the algorithm is given in \cref{fig:solver_overview}.
  
  \subsection{Level 1: \LevelOneLong{}}

  \LevelOneLong{} (\LevelOne{}) computes a valid operating schedule by solving a sequential task assignment problem. The assignment problem is a relaxation of the full \PCTAPF{} problem, because it ignores the constraint that robots must not collide. The solution to the assignment problem is an \emph{optimistic} operating schedule, and its makespan is a lower bound on the makespan of the optimal solution to the full \PCTAPF{} problem. The optimistic schedule is passed to Level 2, \LevelTwo{}, which returns a corresponding minimum-makespan valid route plan. The makespan of this route plan (which may be higher than the lower bound, since \LevelTwo{} must respect the collision-free constraints) constitutes an upper bound on the makespan of the optimal valid solution. \LevelOne{} tries all possible assignments in best-first order until no assignments remain with better makespan than the best route plan found by \LevelTwo{} up to that point. \LevelOne{} is summarized in \cref{alg:level_one}.

  The relaxed problem is formulated as a mixed integer linear program (MILP). We re-cast the sequential assignment problem as a one-off assignment problem by introducing $\numtasks$ ``dummy robots'', where the $j$th dummy robot is shorthand for ``the robot that just delivered object $j$.'' We encode the discrete decision variable as an binary \emph{assignment matrix} $\assignment[]{} \in \booleans^{(\numrobots+\numtasks) \times \numtasks}$, where $\assignment[i]{j} = 1$ indicates that robot $i$ is assigned to transport object $j$. 
  The first $\numrobots$ rows of $\assignment[]{}$ correspond to the real robots, and the final $\numtasks$ rows correspond to the dummy robots. Hence, if $\assignment[i]{j} = 1$ and $\assignment[j+\numrobots,]{k} = 1$, then robot $i$ is assigned to deliver object $j$ and then to deliver object $k$. Because the $j$th dummy robot is a placeholder for the robot that already delivered object $j$, constraints must be added so that dummy robot $j$ cannot be assigned to object $j$ or any of its prerequisites (as this would lead to a cyclic schedule graph). 
   
  To simplify notation, we define $\RobotStartTime[i]$, $\TaskStartTime[j]$, and $\TaskCompletionTime[j]$ as the start time of robot $i$, the pickup time of object $j$, and the delivery time of object $j$, respectively.
  The MILP formulation is given by
  \begin{alignat}{3}
    & \text{minimize}   & \quad & \TaskCompletionTime[\numtasks] \label{eqn:milp_makespan_def} \\
    & & & \assignment[i]{j} \in \{0,1\} \quad i \in 1:\numrobots+\numtasks, j \in 1:\numtasks \label{eqn:milp_binary} \\[-0.3ex] 
    & & & \textstyle \sum_{i=1}^{\numrobots+\numtasks} \assignment[i]{j} = 1, \quad j \in 1:\numtasks \label{eqn:milp_one_robot_per_task} \\[-0.3ex] 
    & & & \textstyle \sum_{j=1}^{\numtasks} \assignment[i]{j} \leq 1, \quad i \in 1: \numrobots+\numtasks \label{eqn:milp_one_task_per_robot} \\[-0.3ex] 
    & & & \assignment[j+\numrobots,]{k} = 0, \quad j \in 1:\numtasks, \quad k \in \predecessors{j} \label{eqn:milp_upstream_constraints} \\[-0.3ex] 
    & & & \RobotStartTime[i] = 0, \quad i \in 1:\numrobots \label{eqn:milp_robot_ics} \\
    & & & \RobotStartTime[j+\numrobots] = \TaskCompletionTime[j], \quad j \in 1:\numtasks  \label{eqn:milp_dummy_starts} \\[-0.3ex] 
    & & & \TaskStartTime[j] \geq 0, \quad j \in 1:\numtasks \label{eqn:milp_object_ics} \\[-0.3ex]
    & & & \TaskStartTime[j] \geq \TaskCompletionTime[k] + \OPERATIONvtxAbbrev{}.\processtime[], \quad \OPERATIONvtxAbbrev{} \in \ProjectSpec.operations, \nonumber \\[-0.3ex] 
    & & & \quad j \in \OPERATIONvtxAbbrev{}.inputs, \ k \in \OPERATIONvtxAbbrev{}.outputs \label{eqn:milp_precedence_constraints} \\ 
    & & & \TaskCompletionTime[j] \geq \TaskStartTime[j] + \COLLECTaction{}_j.\processtime[] + \distmatrix[{\PickupLocation[j]}]{\DropoffLocation[j]} \nonumber \\[-0.3ex] 
    & & & \quad + \DEPOSITaction{}_j.\processtime[], \quad j \in 1:\numtasks \label{eqn:milp_min_task_time} \\[-0.3ex] 
    & & & \TaskStartTime[j] - (\RobotStartTime[i] + \distmatrix[{\RobotPosition[i]{0}}]{\PickupLocation[j]}) \nonumber  \geq -M (1 - \assignment[i]{j}), \nonumber \\[-0.3ex]
    & & & \quad \quad i \in 1:\numrobots, \quad j \in 1:\numtasks,  \label{eqn:milp_big_M} 
  \end{alignat}
  where \eqref{eqn:milp_makespan_def} defines the project makespan, \eqref{eqn:milp_binary} constrains the elements of the assignment matrix to be binary, \eqref{eqn:milp_one_robot_per_task} ensures that each task is assigned to exactly one robot, \eqref{eqn:milp_one_task_per_robot} ensures that each robot (including dummies) is assigned to no more than one task, \eqref{eqn:milp_upstream_constraints} prevents each dummy robot $j$ from being assigned to any task $k \in \predecessors{j}$ among the predecessors of its parent task, \eqref{eqn:milp_dummy_starts} constrains the start time of each dummy robot to match the completion time of its parent task, \eqref{eqn:milp_robot_ics} defines the start times for all non-dummy robots, \eqref{eqn:milp_object_ics} ensures that no object can be collected before time $t=0$, \eqref{eqn:milp_precedence_constraints} enforces all task precedence constraints that arise from the operations in the project spec, \eqref{eqn:milp_upstream_constraints} constrains the time between task start and task completion to be separated by at least collection time plus the minimum travel time between the pick-up and drop-off locations plus the deposit time, and \eqref{eqn:milp_big_M} uses the big $M$ method to constrain the start time of task $j$ to be at least the start time of the assigned robot plus the minimum travel time $\distmatrix[{\RobotPosition[i]{0}}]{\DropoffLocation[j]}$ from the robot initial location to the object pick-up location. The MILP can be solved with any off-the-shelf solver. We use Gurobi in our experiments \cite{gurobi}.

  \begin{algorithm}
    \scriptsize
    \caption{\small \LevelOneLong{} (\LevelOne)}
    \label{alg:level_one}
    \begin{algorithmic}[1]
      \Procedure{\LevelOne}{$\problemInstance$}
        \State $\milp \gets \formulateMILP(\problemInstance)$
        \State $(\bestSolution, \bestCost) \gets (\nullSolution, \infty)$ \Comment best solution 
        \State $\lowerBoundCost \gets 0$ \Comment lower bound 
        \While{$\bestCost > \lowerBoundCost$}
          \State $\ProjectSchedule, \ \lowerBoundCost \gets \solveMILP(\milp)$
          \If{$\bestCost > \lowerBoundCost$}
            \State $\pctapfSolution, \ \solutionCost \gets \textsc{\LevelTwo}(\problemInstance,\ProjectSchedule)$ \Comment Call Level 2
            \If{$\solutionCost < \bestCost$}
              \State $(\bestSolution, \bestCost) \gets (\pctapfSolution, \solutionCost)$
            \EndIf
          \State $\milp.\constraints \gets (\ProjectSchedule \neq \pctapfSolution.\ProjectSchedule)$
          \EndIf
        \EndWhile
        \State \Return $\bestSolution, \bestCost$
      \EndProcedure
    \end{algorithmic}
  \end{algorithm}
  
  \subsection{Level 2: \LevelTwoLong{}}
  
  \LevelTwo{} is a general framework for multi agent path finding \cite{Sharon2012}.
    It operates by best-first search over a binary constraint tree, where each node of the constraint tree contains a set of constraints (the root node constraint set is empty). A \emph{state constraint} $\CBSStateConstraint = \CBSStateConstraintDef[\RobotID]{\RobotPosition[]{}}{t}$ specifies that robot $\RobotID$ may not occupy position $\RobotPosition[]{}$ at time $t$, and an \emph{action constraint} $\CBSActionConstraint = \CBSActionConstraintDef[\RobotID]{\RobotPosition[]{},\RobotState[]{\prime}}{t}$ specifies that robot $\RobotID$ may not move from position $\RobotState[]{}$ to position $\RobotState[]{\prime}$ at time $t$.
    
  When \LevelTwo{} receives a operating schedule from \LevelOne{}, it initializes the root node of the binary constraint tree. 
    Beginning at the root node, \LevelTwo{} passes the operating schedule and the current node's constraint set down to the next level of planner. \LevelThree{} returns a \emph{consistent} route plan that also respects all constraints passed in by \LevelTwo{}, or an infeasible flag if no consistent route plan can be found.
        If the route plan is invalid (i.e., it has at least one conflict), \LevelTwo{} \emph{branches} by generating a pair of mutually exclusive constraints and adding each to a new search node that also inherits all the constraints of its parent search node. \LevelTwo{} calls \LevelThree{} for each child search node to recompute a new solution that respects the added constraint, then adds each search node to a priority queue with priority defined by its makespan. \LevelTwo{} continues until (1) a valid solution is obtained, (2) the cost of the best node in the queue exceeds the upper bound on the optimal solution, or (3) the priority queue is exhausted. In the latter two cases, \LevelTwo{} returns no solution.
  
  \subsection{Level 3: \LevelThreeLong{}}\label{sec:ISPS}
  \begin{algorithm}
    \scriptsize
    \caption{\LevelThreeLong{} (\LevelThree)}\label{alg:level_three}
    \begin{algorithmic}[1]
      \Procedure{\LevelThree}{$\ProjectSchedule,\constraints$}
        \State $\RoutePlan \gets ([\RobotPosition[1]{0}],\ldots,[\RobotPosition[\numrobots]{0}])$ \Comment init route plan
        \State $\ClosedSet \gets \emptyset$ \Comment closed set
        \State $\NodeQueue \gets \textsc{InitQueue}(\ProjectSchedule,\ClosedSet)$ \Comment priority queue \label{eqn:init_node_queue}
        \While{$\NodeQueue \neq \emptyset$}
          \State $\ScheduleNode \gets \textsc{dequeue}(\NodeQueue)$ \Comment pop vertex from queue
          \State $\AgentPath{} \gets \textsc{\LevelFour}(\RoutePlan,\ScheduleNode,\constraints)$ \Comment plan path segment $\AgentPath{}$
          \State $\vtx.\CompletionTime[] \gets \AgentPath{}.\CompletionTime[]$ \Comment update vertex final time
          \State $\RoutePlan \gets \addPathToRoutePlan(\RoutePlan,\AgentPath{})$ \Comment update route plan
          \State $\ClosedSet \gets \{\ScheduleNode\} \cup \ClosedSet$ \Comment update closed set
          \State $\ProjectSchedule \gets \textsc{UpdateSchedule}(\ProjectSchedule)$  \label{eqn:update_schedule}
          \State $\NodeQueue \gets \textsc{InitQueue}(\ProjectSchedule,\ClosedSet)$  \label{eqn:update_schedule}
        \EndWhile
        \State \Return $\RoutePlan$
      \EndProcedure
      \Statex
      \Procedure{UpdateSchedule}{$\ProjectSchedule$}
        \For{$\vtx \in \textsc{topological\_sort}(\ProjectSchedule)$}
          \For{$\vtxTWO \in \predecessors{\vtx}$}
            \State $\vtx.\StartTime[] \gets \max(\vtx.\StartTime[],\vtxTWO.\FinalTime)$
          \EndFor
          \State $\vtx.\CompletionTime[] \gets \max(\vtx.\StartTime[] + \vtx.\processtime[],\vtx.\CompletionTime[])$
        \EndFor
        \For{$\vtx \in \textsc{reverse\_topological\_sort}(\ProjectSchedule)$}
          \For{$\vtxTWO \in \successors{\vtx}$}
            \State $\vtx.\Slack \gets \min(\vtx.\Slack,\vtxTWO.\StartTime[] + \vtxTWO.\Slack - \vtx.\CompletionTime[])$
          \EndFor
          \State $\vtx.\CompletionTime[] \gets \max(\vtx.\StartTime[] + \vtx.\processtime[],\vtx.\CompletionTime[])$
        \EndFor
        \State \Return $\ProjectSchedule$
      \EndProcedure
      \Statex
      \Procedure{InitQueue}{$\ProjectSchedule,\ClosedSet$}
        \For{$\vtx \in \Vertices(\ProjectSchedule)$}
            \If{$\ScheduleNode \notin \ClosedSet \land \predecessors{\ScheduleNode} \subseteq \ClosedSet$}
            \State $\NodeQueue \gets \textsc{enqueue}(\NodeQueue, \ScheduleNode \rightarrow \ScheduleNode.\Slack)$
            \EndIf
        \EndFor
        \State \Return $\NodeQueue$
      \EndProcedure
    \end{algorithmic}
\end{algorithm}
  
  The \LevelThreeLong{} (\LevelThree{}) module receives a operating schedule and a set of constraints from CBS.
    \LevelThree{} incrementally constructs a consistent plan by traversing the schedule graph in topological order and calling \LevelFour{} to compute a path segment for each vertex.
      The key idea behind \LevelThree{} is that it takes advantage of \emph{slack} in the operating schedule.
    Slack denotes the amount of room for delay in a vertex's final time $\vtx.\FinalTime$ before the makespan would increase. Vertices with high slack can afford significant delay without affecting the project completion time. Vertices with zero slack are on the \emph{critical path}. If they are delayed by even a single timestep, the entire project will be delayed.
  
  \LevelThree{} begins by adding all root vertices (vertices with no predecessors) of the schedule graph to a priority queue $\NodeQueue$ where they are prioritized by their slack. The lowest slack vertex is popped from the queue, and \LevelFour{} is called to plan a path segment corresponding to that vertex. When planning is complete for that vertex, it is placed in the \emph{closed set} $\ClosedSet$. The path segment is added to the partial route plan, and \LevelThree{} calls a subroutine to update the operating schedule and recompute the slack of all vertices.
  When all of a vertex's predecessors are in the closed set, that vertex is added to the priority queue.
  
  Only \GOaction{} and \CARRYaction{} nodes require actual path-planning from an initial location to a target location.\footnote{\GOaction{} and \CARRYaction{} are functionally equivalent--the path-planning process for them is identical. We simply call them different things to acknowledge that the agent is actually transporting an object during the \CARRYaction{} phase.} All other nodes in the operating schedule serve only as checkpoints, and can be automatically closed as soon as they become active. To avoid scheduling explicit waiting periods between \GOaction{} nodes and \COLLECTaction{} nodes, the planner extends the planning horizon of each \GOaction{} node to the time step at which the assigned object become available. \LevelThree{} terminates when either (1) all schedule nodes are closed, (2) \LevelFour{} fails to find a feasible path, or (3) the cost of the route plan exceeds the upper bound cost maintained by \LevelOne{}. \LevelThree{} is summarized in \cref{alg:level_three}.

  \subsection{Level 4: \LevelFourLong{}}
  
  \LevelFourLong{} (\LevelFour{}) combines a custom cascading search heuristic with the well-known A$^\star$ graph-search algorithm \cite{Hart1968}. As in regular space-time A$^\star$, \LevelFour{} maintains a priority queue of search states, where each search state corresponds to a path $\AgentPath{}$. At every iteration, \LevelFour{} pops the lowest-cost search state from the queue and expands it by trying all possible one-step actions beginning from the terminal state of the corresponding path. When a search state satisfies the termination criteria (i.e., the path reaches the goal location), it is returned to \LevelFour{}.

  For brevity, we do not include the pseudocode of the entire \LevelFour{} algorithm. However, \cref{alg:A_star_cost} shows the pseudocode of $\textsc{get\_heuristic\_cost}(\RoutePlan,\AgentPath{},\vtx)$, a subroutine that computes the four element heuristic cost $\PathCost \in \reals_+^4$ of a path $\AgentPath{}$. This custom heuristic cost determines the order in which \LevelFour{} will expand search states within the algorithm. Ties are broken in cascading fashion. This cascaded tie-breaking behavior is the key to the efficiency of our path-planner in computing make-span optimal route plans while simultaneously avoiding conflicts.

  The first element $\PathCost_1$ of the cost tuple is the \emph{delay} cost. It measures the amount by which a path is guaranteed to delay the entire project, which is $0$ until the slack disappears. 
  The second cost element $\PathCost_2$ counts the number of times that a path conflicts with the global route plan $\RoutePlan$. 
  The third element $\PathCost_3$ is equal to the path length plus the remaining distance to the goal location, and the fourth element $\PathCost_4$ is simply the distance to the goal location.
  Hence, the default behavior (based on $\PathCost_3$ and $\PathCost_4$) of \LevelFour{} is to move to the goal location as quickly, but it will avoid conflicts at the expense of path length until the slack runs out, at which time path length again become the dominant criterion for optimality. The only way that \LevelFour{} will return a path that creates conflicts is if it runs out of slack and there is no other path of equal or shorter length with fewer conflicts.


  %
  
\floatstyle{spaceruled}
\restylefloat{algorithm}
  \begin{algorithm}
  \scriptsize
  \caption{\footnotesize \textsc{get\_heuristic\_cost} Subroutine of \LevelFour{}}
  \label{alg:A_star_cost}
  \begin{algorithmic}[1]
    \Procedure{get\_heuristic\_cost}{$\AgentPath{},\RoutePlan,\vtx$} \label{alg:get_cost}
      \State $\HeuristicCost \gets \distmatrix[\pathstate]{\vtx.goal\_state}$ \Comment ``cost-to-go'' heuristic \label{alg:dist_function}
      \State $\PathCost_1 \gets \max (0, \AgentPath{}.\CompletionTime[] + \HeuristicCost - (\vtx.\CompletionTime[] + \vtx.\Slack))$ \Comment delay cost \label{alg:slack_cost}
      \State $\PathCost_2 \gets \CountConflicts(\AgentPath{},\RoutePlan)$ \Comment number of conflicts along path
      \State $\PathCost_3 \gets \AgentPath{}.length + \HeuristicCost$ \Comment lower bound total travel distance
      \State $\PathCost_4 \gets \HeuristicCost$ \Comment minimum remaining travel time
      \State \Return $\PathCost$
    \EndProcedure
  \end{algorithmic}
  \end{algorithm}

  \subsection{Repairing Route Plans}
  
  When \LevelThree{} has finished computing a consistent route plan for the entire operating schedule, the solution is checked for conflicts. If there are conflicts, \LevelThree{} tries to repair the ``broken'' solution by iterating through the operating schedule a second time. The difference the second time through is that a full route plan (as opposed to a partial route plan) is available to populate the conflict table which is used in \LevelFour{} for computing the conflict cost $\CountConflicts(\AgentPath{},\RoutePlan)$. This allows paths that are planned early in the topological ordering to be adjusted to avoid potential conflicts with plans that are planned later in the \LevelThree{} procedure.

  \subsection{Theoretical Properties}

  It can be shown that \LevelOne{} will search the full space of possible valid project schedules in best-first order. If the combination of \LevelTwo{}, \LevelThree{}, and \LevelFour{} is optimal and complete, the full algorithm can also be guaranteed to find the optimal solution if the problem is feasible. \citeauthor{Sharon2012} have already shown that \LevelTwo{} is optimal and complete under the assumption that the lower level path planner is also optimal and complete \cite{Sharon2012}. 

  However, our lower level path planner (\LevelThree{} combined with \LevelFour{}) is incomplete because it is possible for \LevelFour{} to prematurely consume slack in one vertex such that a downstream vertex is forced to unnecessarily delay the project. Completeness could be ensured by removing the slack term from \cref{alg:slack_cost} of \cref{alg:A_star_cost}, but this would negate the planner's key advantage: its ability to avoid conflicts by consuming slack. We are preparing an extension of \LevelThree{} and \LevelFour{} that will incrementally anneal the slack cost term when necessary, thus achieving completeness without ruining the planner's efficiency.
  
  \section{Experiments}
\begin{figure}[ht]
  \centering
  \subcaptionbox{Runtime statistics for the full \PCTAPF{} solver. \label{fig:full_alg_results}}{%
      \input{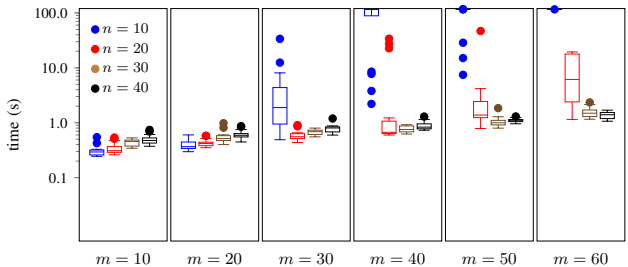}%
    }\par\medskip
  \subcaptionbox{Runtime statistics for the task assignment MILP solver. \label{fig:assignment_results}}{%
      \input{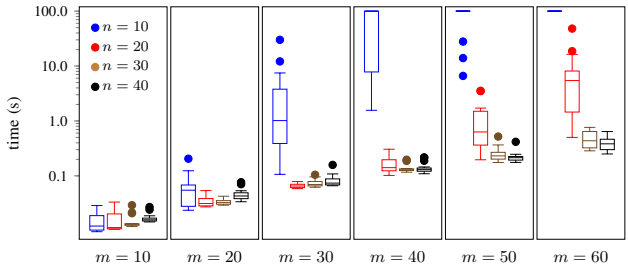}%
    }\par\medskip        
  \subcaptionbox{Runtime statistics for \LevelThree{}. \label{fig:low_level_results}}{%
      \input{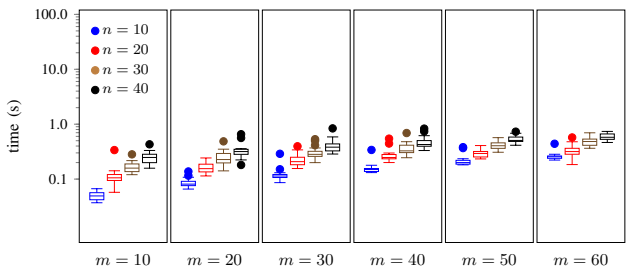}%
    }
  \caption{Box plots summarizing the computation time for the full \PCTAPF{} solver, the \LevelOne{} MILP Solver and \LevelThree{}. Each individual plot represents 16 trials with the indicated values of $\numtasks$ (labeled along the $x$-axis) and $\numrobots$ (increasing from left-to-right and labeled by color). Runtime is plotted on a log scale, and all three figures are plotted on the same scale.}
  \label{fig:results}
  \end{figure}

  We evaluate our solver on a set of \numproblems{} randomly generated \PCTAPF{} problem instances---16 instances for every combination of number of robots $\numrobots \in \{10,20,30,40\}$ and number of objects $\numtasks \in \{10,20,30,40,50,60\}$. 
  The grid world environment is shown in \cref{fig:factory_env}. Initial robot locations are selected at random. The initial and destination locations of all objects are selected at random from the designated pick-up and drop-off zones surrounding each manufacturing station. For each problem instance, we evaluate the runtime of (a) the full algorithm, (b) the \LevelOne{} MILP solver alone, and (c) a single iteration of \LevelThree{}. The results are summarized in \cref{fig:full_alg_results,fig:assignment_results,fig:low_level_results}.
  
  \Cref{fig:full_alg_results} shows that the \PCTAPF{} solver runs very fast for most problem instances, even when $\numtasks$ and $\numrobots$ are large. However, \LevelOne{} struggles when the ratio of tasks to robots becomes high. This is unsurprising, as the MILP approaches a traveling salesman problem when $\numtasks \gg \numrobots$. The \LevelOne{} MILP solver timed out (at 100 seconds) on 40 of the problem instances (including all instances with $\numtasks=60$, $\numrobots=10$) before finding an optimal solution. In these cases, the suboptimal MILP solution was still passed to \LevelTwo{}.
  
\begin{figure}[ht]
  \vspace{5pt}
  \centering
  \input{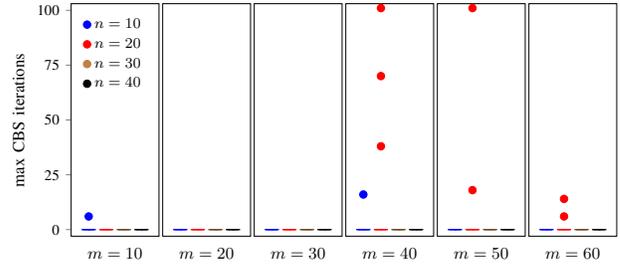}%
  \caption{Box plots summarizing the maximum number of \LevelTwo{} iterations in any full iteration of the \PCTAPF{} solver. Plots are organized as in \cref{fig:full_alg_results,fig:assignment_results,fig:low_level_results}. Only nine of the 384 problem instances required any \LevelTwo{} branching.}
  \label{fig:cbs_results}
  \end{figure}

  \LevelThree{} (as demonstrated by \cref{fig:low_level_results}) scales gracefully with problem size. The route planner's efficiency is largely due to the ability of \LevelThree{} and \LevelFour{} to exploit slack in the operating schedule, thereby removing the burden of conflict-resolution from \LevelTwo{}. \Cref{fig:cbs_results} shows that only nine of the 384 problem instances required any \LevelTwo{} branching at all. On two problem instances, the solver reached our arbitrary limit of 100 branching operations in a single \LevelTwo{} iteration.  
  Despite the incompleteness of \LevelThree{} and \LevelFour{}, only three problem instances (besides those on which the solver reached a time or iteration limit) were not solved optimally.

  \newcommand{\plotarraywidth}{3.5cm}
  \newcommand{\plotarrayheight}{6cm}
  \renewcommand{\plotarraywidth}{3.25cm}
  \renewcommand{\plotarrayheight}{6.0cm}

  \renewcommand{\plotarraywidth}{3.25cm}
  \renewcommand{\plotarrayheight}{6.0cm}
  
  In addition to the experiments above, we include a video to demonstrate our solver's performance in a high-fidelity simulator with multiple non-holonomic differential drive robots.
      After solving the \PCTAPF{} problem, we use geometric path-planning and convex optimization to convert the optimal route plan into dynamically feasible trajectories that allow each robot to pass in and out of each grid cell in the planned sequence at precise intervals.
          To stabilize each robot about its reference trajectory, we implement a hybrid nonlinear feedback control policy that switches between control laws (including the tracking control law of \citeauthor{Lee2001} \cite{Lee2001}) depending on the current stage of the reference trajectory.
      Our simulator is built on Webots (\url{http://www.cyberbotics.com}).
  
  \section{Conclusions}
  We introduced the \PCTAPF{} formulation and presented a hierarchical algorithm that takes advantage of slack in the operating schedule to efficiently and optimally solve many \PCTAPF{} problem instances. 
  Though the algorithm performs well empirically, there are several weaknesses that need to be addressed.
  \LevelThree{} (and hence, the entire algorithm) is not complete. We are preparing an extension of \LevelThree{} that will make the full algorithm optimal and complete.
  Moreover, we have observed that the computational cost of \LevelOne{} can grow intractably large for certain classes of problems. This makes it attractive to develop bounded-suboptimal \PCTAPF{} solvers with better runtime guarantees.

  Real manufacturing applications are likely to include large sub-assemblies that are too big to be transported by a single robot.
    We are currently extending our approach to ``collaborative transport'' scenarios, in which some transport tasks must be handled by teams of robots.
      Another interesting avenue of future work would be to extend our work to a heterogeneous robot fleets, wherein robots might vary in size, shape, speed, and ability to service certain tasks.

  We are extending our algorithm to a paradigm where projects arrive at the factory continuously, and wherein better performance may be attained by partial re-planning of the operating schedule and route plan to optimally balance multiple projects simultaneously.
    A related theme in our ongoing work is robustness to uncertainty. We are interested in incorporating open-loop robustness to potential failure and delay, as well as closed-loop robustness during execution of the \PCTAPF{} solution. Our algorithm currently assumes a deterministic environment, but several efforts from the MAPF and TAPF literature offer a promising starting point for incorporating robustness to uncertainty \cite{Honig}, \cite{Ma2017a}.



\renewcommand*{\bibfont}{\small}
\printbibliography

\end{document}